\pdfoutput=1

\documentclass[11pt]{article}

\usepackage{naacl2021}

\usepackage{times}
\usepackage{latexsym}
\usepackage{enumitem}
\usepackage{bm}
\usepackage{amsmath}
\usepackage{verbatim}
\usepackage{graphicx}
\usepackage[T1]{fontenc}

\usepackage[utf8]{inputenc}

\usepackage{microtype}

\title{GTN-ED: Event Detection Using Graph Transformer Networks}

\author{Sanghamitra Dutta\thanks{*Research work was done while Sanghamitra Dutta was interning at Dataminr Inc. } \\
  Carnegie Mellon University \\
  \texttt{sanghamd@andrew.cmu.edu} \\\And
  Liang Ma\\
  Dataminr \\
  \texttt{lma@dataminr.com} \\\And
  Tanay Kumar Saha\\
  Dataminr \\
  \texttt{tsaha@dataminr.com} \\ \AND
  Di Lu\\
  Dataminr \\
  \texttt{dlu@dataminr.com} \\\And
  Joel Tetreault\\
  Dataminr \\
  \texttt{jtetreault@dataminr.com } \\\And
  Alejandro Jaimes\\
  Dataminr \\
  \texttt{            ajaimes@dataminr.com}}

\begin{document}
\maketitle
\begin{abstract}
Recent works show that the graph structure of sentences, generated from dependency parsers, has potential for improving event detection. However, they often only leverage the edges (dependencies) between words, and discard the dependency labels (e.g., nominal-subject), treating the underlying graph edges as homogeneous. In this work, we propose a novel framework  for incorporating both dependencies and their labels using a recently proposed technique called Graph Transformer Networks (GTN). We integrate GTNs to leverage dependency relations on two existing homogeneous-graph-based models, and demonstrate an improvement in the F1 score on the ACE dataset.
\end{abstract}

\section{Introduction}
Event detection is an important task in natural language processing, which encompasses predicting important incidents in texts, e.g., news, tweets, messages, and manuscripts~\cite{yang2016joint,jrnn,feng-etal-2016-language,zhang2020question,du2020event,mcclosky-etal-2011-event-extraction,ji-grishman-2008-refining,liao2010using,li2013joint,yang2019exploring}. As an example, consider the following sentence: \textit{The plane arrived back to base safely.} Here, the word \emph{arrived} is an event trigger that denotes an event of the type ``Movement:Transport,'' while ``\emph{The plane}'' and ``\emph{base}'' are its arguments. Given a sentence, the objective of the event detection task is to predict all such event triggers and their respective types. 

\begin{figure}
    \centering
    \includegraphics[width=3cm]{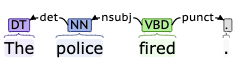} \hspace{0.2cm}
    \includegraphics[width=2.8cm]{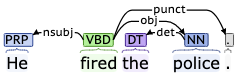}
    \caption{Examples of syntactic dependency parsing. }
    \label{fig:example}
\end{figure}

Recent works on event detection~\cite{gcnAAAI,liu-etal-2018-jointly,moganed,sdp} employ graph based methods (Graph Convolution Networks~\cite{gcn_org}) using the dependency graph (shown in Fig.~\ref{fig:example}) generated from syntactic dependency-parsers. These methods are able to capture useful non-local dependencies between words that are relevant for event detection. However, in most of these works (with the notable exception of \citet{relationaware}), the graph is treated as a homogeneous graph, and the dependency labels (i.e., edge-types in the graph) are ignored.

Dependency labels can often better inform whether a word is a trigger or not. Consider the two sentences in Fig.~\ref{fig:example}. In both the sentences, there is an edge between ``\textit{police}'' and ``\textit{fired}''. A model that does not take into account dependency labels will only have access to the information that they are connected. However, in the first sentence,  ``\textit{fired}'' is an event trigger of type ``Conflict:Attack,'' whereas in the second sentence, it is of type ``Personnel:End Position.'' The fact that the edge label between ``\textit{police}'' and ``\textit{fired}'' is a nominal-subject or an object relation serves as an indicator of the type of event trigger. Hence, leveraging the dependency labels can help improve the event detection performance.

In this work, we propose a simple method to employ the dependency labels into existing models inspired from a recently proposed technique called Graph Transformer Networks (GTN)~\cite{gtn}. GTNs enable us to learn a soft selection of edge-types and composite relations (e.g., multi-hop connections, called \emph{meta-paths}) among the words, thus producing heterogeneous adjacency matrices. 

We integrate GTNs into two homogeneous-graph-based models (that previously ignored the dependency relations), namely, a simple gated-graph-convolution-based model inspired by \citet{gcnAAAI,liu-etal-2018-jointly,sdp}, and the near-state-of-the-art MOGANED model~\cite{moganed}, enabling them to now leverage the dependency relations as well. Our method demonstrates a relative improvement in the F1 score on the ACE dataset~\cite{ace} for both models, proving the value of leveraging dependency relations for a graph-based model. While the goal of this paper is not to establish a state-of-the-art (SOTA) method, but rather to show the merit of our approach, we do note that the improvements with our method approach the current SOTA~\cite{relationaware} (which leverages dependency relations using embeddings instead of GTNs).

To summarize, our main contribution is \textit{a method of enabling existing homogeneous-graph-based models to exploit dependency labels for event detection, inspired from GTNs.} Incorporating GTNs in NLP tasks has received less attention (also see recent related work \citet{veyseh2020graph}).

\noindent \textbf{Notations:} We denote matrices and vectors in bold, e.g., $\bm{A}$ (matrix) or $\bm{a}$ (vector). Note that, $A(u,v)$ denotes the element at index $(u,v)$ in matrix $\bm{A}$. 


\begin{figure}
    \centering
    \includegraphics[height=2.2cm]{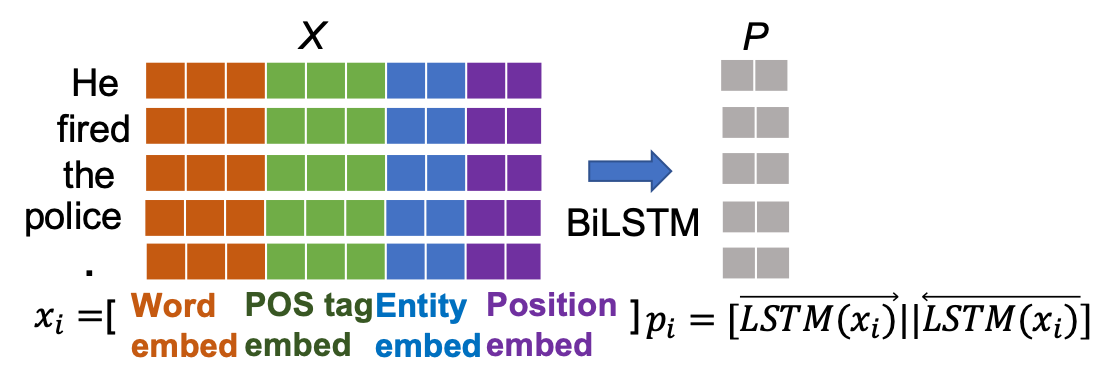}
    \caption{Embedding and BiLSTM Module.}
    \label{fig:embedding}
\end{figure}
\section{Proposed Method}
\label{sec:method}
In this work, we incorporate GTNs onto two homogeneous-graph-based models: (i) {\bf Model I}: a gated-graph-convolution-based model inspired by~\citet{gcnAAAI,liu-etal-2018-jointly,sdp}; and (ii) {\bf Model II}: MOGANED model~\cite{moganed}. Both models have a similar initial embedding and BiLSTM module, followed by a graph-based module (where their differences lie), and finally a classification module.

\begin{figure}
    \centering
    \includegraphics[height=2.9cm]{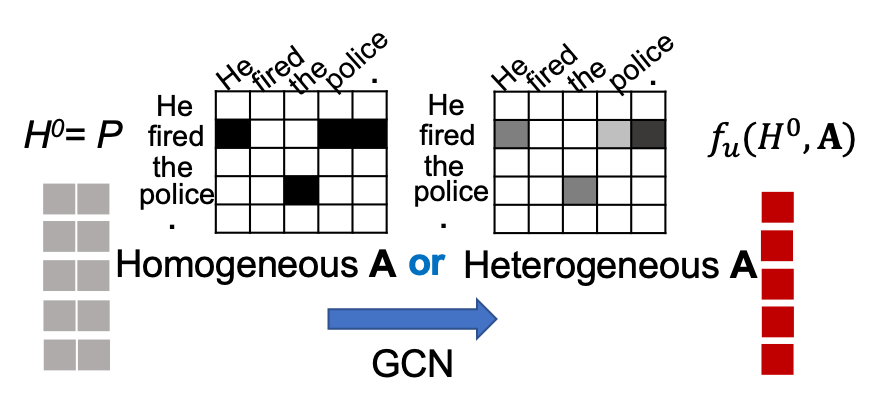}
    \caption{Basic Gated-Graph-Convolution Network.}
    \label{fig:gcn}
\end{figure}

\noindent \textbf{Embedding and BiLSTM Module:} Our initial module (shown in Fig.~\ref{fig:embedding}) is similar to existing works (e.g., \citet{moganed}). Each word of the sentence is represented by a token which consists of the word embedding, the POS tag embedding, the Named-Entity type embedding, and its positional embedding. For a sentence of $n$ words, we denote this sequence of tokens as $X=x_0,x_1, \ldots,x_{n-1}$. Next, we introduce a
BiLSTM to encode $X$ into its context $P=p_0,p_1,\ldots,p_{n-1}$ where $p_i=[\overset{\rightarrow}{LSTM}(x_i)||\overset{\leftarrow}{LSTM}(x_i)]$, and $||$ denotes the concatenation operation.  $P$ is then fed to the graph-based module, as discussed next.

\noindent \textbf{Graph-Based Module:} We first introduce the basic unit of both Model I and II, i.e., gated-graph-convolution network (see Fig.~\ref{fig:gcn}). Let $H^k=h_0^k,h_1^k,\ldots,h_{n-1}^k$ be the input and $H^{k+1}=h_0^{k+1},h_1^{k+1},\ldots,h_{n-1}^{k+1}$ be the output of the $k$-th layer of this module with $H^0=P$. Given any adjacency matrix $\bm{A}$ and input $H^k$, consider the following operation at layer $k$: 
\begin{equation}
    f_u(H^k,\bm{A})=\sum_{v=0}^{n-1} G_{\bm{A}}^k(u,v)(\bm{W}^k_{\bm{A}}h^k_v+\bm{b}^k_{\bm{A}}). \label{eq:gcn} 
\end{equation}
Here, $\bm{W}^k_{\bm{A}}$ and $\bm{b}^k_{\bm{A}}$ are the weight matrix and bias item for the adjacency matrix $\bm{A}$ at layer $k$, and $G_{\bm{A}}^k(u,v)$ is the  gated-importance, given by $G_{\bm{A}}^k(u,v)=A(u,v)\sigma(\bm{w}^k_{att,\bm{A}}h^k_v+\epsilon_{att,\bm{A}}^k)$, where $\sigma(\cdot)$ is an activation function, and $\bm{w}_{att,\bm{A}}$ and $\epsilon_{att,\bm{A}}$ are the attention weight vector and bias item.
\begin{figure}
    \includegraphics[width=3.8cm]{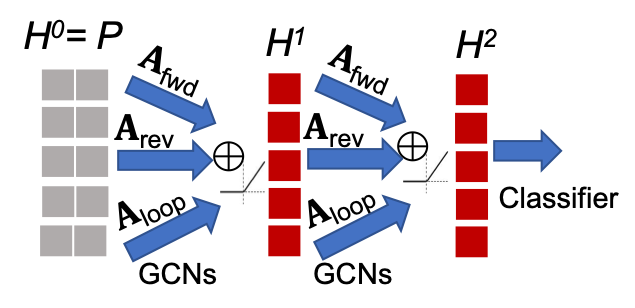}
    \includegraphics[width=3.8cm]{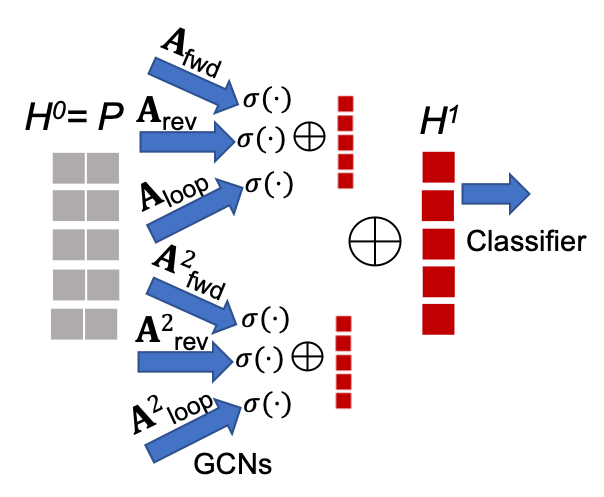}
    \caption{(Left) Model I; (Right) Model II.}
    \label{fig:models}
\end{figure}

A dependency parser, e.g., Stanford Core NLP~\cite{manning-etal-2014-stanford}, generates a directed heterogeneous graph $\mathcal{G}$ for each sentence (recall Fig.~\ref{fig:example}). Existing works typically do not use the dependency labels (e.g.,  nominal-subject); they only derive three homogeneous adjacency matrices from $\mathcal{G}$ as follows: (i) $\bm{A}_{fwd}$ where $A_{fwd}(i,j)=1$ if there is an edge from node $i$ to $j$; (ii) $\bm{A}_{rev}$  where $A_{rev}(i,j)=1$ if there is an edge from node $j$ to $i$; and (iii) $\bm{A}_{loop}$ which is an identity matrix. 

For \textbf{Model I} (see Fig.~\ref{fig:models} (Left)), the output of the $k$-th layer (input to $k+1$-th layer) is given by
$h^{k+1}_u{=}\mathrm{ReLu}(\sum_{\bm{A}\in \{\bm{A}_{fwd},\bm{A}_{rev},\bm{A}_{loop} \}}f_u(H^k,\bm{A})).$ The first layer of gated-graph-convolution network captures dependencies between immediate neighbors ($1$-hop). To capture $K$-hop dependencies, Model I has $K$ consecutive layers of such gated-graph-convolution networks. The output of this graph-based module is then fed to a multi-layer perceptron (MLP) with attention weights for classifying each word into its event-type (or ``not an event'').

In \textbf{Model II}, instead of passing the BiLSTM output $P$ through a \textit{series} of $K$ consecutive gated-graph-convolution layers (to capture $K$-hop connections), this model separately aggregates the outputs of $T$ \textit{parallel} graph-convolution layers with separate adjacency matrices representing hops of length $1,2,\ldots,T$ (see Fig.~\ref{fig:models} (Right)). Let $H^0(=P)$ be the input and $H^1$ be the output of the graph-based module of Model II (which effectively has only one layer, i.e., $k{=}0$). In \citet{moganed}, the homogeneous adjacency matrices $\bm{A}_{fwd}$, $\bm{A}_{rev}$, and $\bm{A}_{loop}$ are considered with their corresponding $t$-hop adjacency matrices $\bm{A}_{fwd}^t$, $\bm{A}_{rev}^t$, and $\bm{A}_{loop}^t$ (multiplied $t$ times) respectively. 
The output of the graph-based module is given by:
$
    h^1_u{=}\sum_{t=0}^{T-1} \bm{w}_{att,t}\bm{v}_t$ where $\bm{v}_t=\sum_{\bm{A}\in \{\bm{A}_{fwd},\bm{A}_{rev},\bm{A}_{loop} \}} \sigma(f_u(H^0,\bm{A}^t)).
$ 
Here, $\bm{w}_{att,t}$ is an attention-weight (further details in \cite{moganed}) and $\sigma(\cdot)$ is the exponential linear unit\footnote{The gated-importance $G^k_{\bm{A}}(u,v)$ has subtle differences between Model I and II.}. Finally, these outputs are passed through an MLP with attention weights for classification.

\noindent \textbf{Remark.} 
The reason for using only three matrices instead of a separate adjacency matrix for each edge-type 
is that it results in an explosion of parameters for the gated-graph-convolution network, as individual weight matrices have to be learnt for each type of edge (see also \citet{gcnAAAI}). In this work, we replace the homogeneous matrices $\bm{A}_{fwd}$ and $\bm{A}_{rev}$ with heterogeneous adjacency matrices without a significant overhead in the number of parameters, as discussed next.
\begin{figure}
    \includegraphics[width=7.6cm]{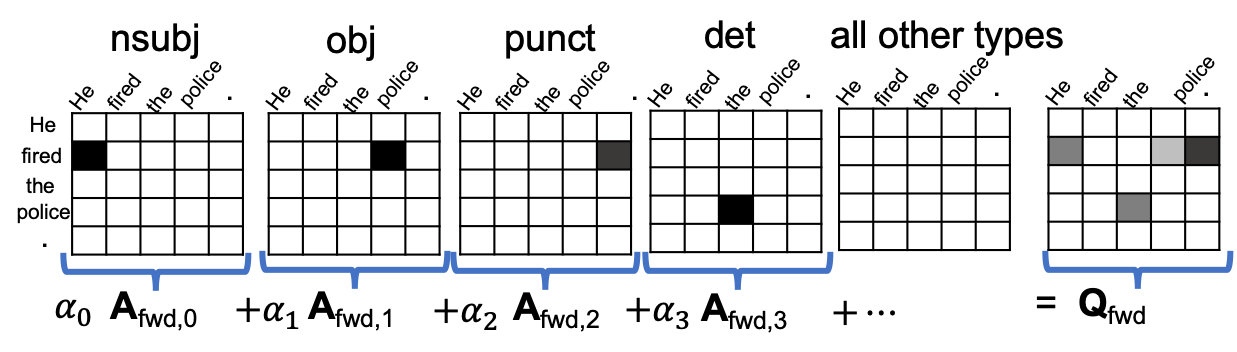}
    \caption{GTN to obtain heterogeneous adjacency matrix of meta-path length $1$ (Recall Fig.~\ref{fig:example} for the graph).}
    \label{fig:gtn}
\end{figure}

\noindent \textbf{Obtaining Heterogeneous Adjacency Matrices With GTN:}
Consider a directed heterogeneous graph $\mathcal{G}$ with each edge belonging to one of $L$ types. This graph can be represented using a set of $L$ adjacency matrices $\{\bm{A}_{fwd,0},\bm{A}_{fwd,1},\ldots,\bm{A}_{fwd, L-1}\}$, each corresponding to a different edge-type (dependency label). $A_{fwd,l}(i,j)=1$ if there is a directed edge from node $i$ to $j$ of type $l$. A GTN obtains a heterogeneous adjacency matrix by learning a convex combination $\bm{Q}_{fwd}=\sum_{l=0}^{L-1}\alpha_l \bm{A}_{fwd,l}$ (see Fig.~\ref{fig:gtn}) where $\bm{\alpha}=\text{softmax}(\bm{w})$ and $\bm{w}$ is a weight vector that the model learns. The matrix $\bm{Q}_{fwd}$ is a heterogeneous adjacency matrix with an ``appropriately weighted'' edge between any two nodes that have an edge in any of the $L$ original matrices. 

For \textbf{Model I}, we first generate a set of $L$ adjacency matrices (for $L$ edge-types) corresponding to the directed forward edges, and another set of $L$ adjacency matrices corresponding to the reverse edges. Next, we learn heterogeneous adjacency matrices, i.e., $\bm{A}_{fwd}=\bm{Q}_{fwd}$ and $\bm{A}_{rev}=\bm{Q}_{rev}$. Our technique enables baseline Model I to leverage dependency relations by learning only $2L$ more scalar parameters which is significantly less than learning individual weight matrices for $L$ edge-types.

For \textbf{Model II}, we not only aim to learn heterogeneous adjacency matrices to replace the homogeneous $\bm{A}_{fwd}$ and $\bm{A}_{rev}$, but also learn heterogeneous adjacency matrices that have an ``appropriately weighted'' edge between every two nodes that are $t$-hops apart in the original graph $\mathcal{G}$ (called a meta-path of length $t$) so as to replace  $\bm{A}_{fwd}^t$ and $\bm{A}^t_{rev}$. 
Specifically, for the case of $t=2$, GTN first learns two convex combinations $\bm{Q}_{fwd,0}$ and $\bm{Q}_{fwd,1}$ (each corresponds to meta-paths of length $1$), and then computes the product $\bm{Q}_{fwd,0}\bm{Q}_{fwd,1}$. Similarly, one can compute a product of $t$ such adjacency matrices to learn meta-paths of length $t$. 

We replace all $t$-hop adjacency matrices with heterogeneous adjacency matrices of meta-path length $t$, learnt through GTNs, e.g., $\bm{A}_{fwd}^t$ is replaced by
$\bm{Q}_{fwd,0}\bm{Q}_{fwd,1}\ldots\bm{Q}_{fwd,t-1}$, where each $\bm{Q}_{fwd,i}$ is a convex combination of $L$-adjacency matrices corresponding to the directed forward edges. Similar heterogeneous adjacency matrices of meta-path length $t$ are learnt for the reverse edges as well to replace $\bm{A}_{rev}^t$. This modification enables the baseline Model II to leverage the dependency relations, by only learning $2Lt$ more scalar parameters for each $t$, which is practicable.

\section{Results}

\textbf{Dataset and Evaluation Metrics:} We use the benchmark ACE2005 English dataset~\cite{ace} with the same data split as in prior works (where the sentences from $529$, $30$, and $40$ documents are used as the training, validation, and test set). We use the Stanford CoreNLP toolkit~\cite{manning-etal-2014-stanford} for sentence splitting, tokenizing, POS-tagging and dependency parsing. We use word embeddings trained over the New York Times corpus with Skip-gram algorithm following existing works~\cite{moganed}. We evaluate the Precision (P), Recall (R) and F1 score.

\noindent\textbf{Model Settings:}
For Model I,  the number of consecutive layers of gated-graph-convolution networks ($K$) is varied from $1$ to $3$. 
For Model II, we use the code\footnote{\url{https://github.com/ll0iecas/MOGANED}} with same hyper parameter settings.



\noindent \textbf{Performance:} For both Models I and II, GTNs demonstrate an improvement of about $1$ point F1 score (see Tables~\ref{tab:gcn} and \ref{tab:moganed}). 
The 76.8 F1 score for Model II with GTNs is also quite close to the SOTA performance of 77.6 for this task~\cite{relationaware}.

\begin{table}
\centering
\scalebox{0.9}{\begin{tabular}{llll}
\hline
\textbf{$K$} & \textbf{P} & \textbf{R} & \textbf{F1}\\
\hline
$1$ & 72.0 &75.5 & 73.7 \\ 
$2$ & 70.7 & 75.7& 73.1 \\
$3$ & 72.8& 70.7& 71.7\\ \hline
\end{tabular}
\begin{tabular}{llll}
\hline
\textbf{$K$} & \textbf{P} & \textbf{R} & \textbf{F1}\\
\hline
$1$ & 72.9 & \textbf{76.4} & \textbf{74.6} \\ 
$2$ & 72.1 & 75.9 & 74.0 \\
$3$ & \textbf{73.8} & 73.6 & 73.7 \\ \hline
\end{tabular}}
\caption{Performance of gated-graph-convolution-based models (\textbf{Model I}) for varying number of consecutive convolution layers ($K$): (Left) Baseline models with no GTNs; (Right) Proposed models with GTNs.  }\label{tab:gcn}
\end{table}

\begin{table}
\centering
\begin{tabular}{lccc}
\hline
\textbf{Method} & \textbf{P} & \textbf{R} & \textbf{F1}\\
\hline
Baseline (no GTNs) & 79.5 & 72.3 & 75.7\\ 
Proposed (with GTNs) & \textbf{80.9} & \textbf{73.2} & \textbf{76.8}\\ \hline
\end{tabular}
\caption{Performance of MOGANED (\textbf{Model II}).}\label{tab:moganed}
\end{table}

\noindent\textbf{Examining Specific Predictions For Insights:} To explain the role of GTNs, we examined all the predictions on the validation set using the baseline Model II (no GTNs) and the proposed Model II (with GTNs). We include some specific instances here that we found interesting and insightful. 

We observe that using GTNs makes the predictions more ``precise,'' by reducing the number of false-positive event trigger detections. For instance, \textit{He's now national director of Win Without War, and \textbf{former} Congressman Bob Dornan, Republican of California.} Here, ``\textit{former}'' is the only event trigger (type Personnel:End-Position), as is correctly identified by our model. However, the baseline model also falsely identifies ``\textit{War},'' as an event trigger of type Conflict:Attack.
Another example is: \textit{In a monstrous conflict of interest, [...].} Here, the baseline falsely identifies ``\textit{conflict},'' as a trigger of type Conflict:Attack. Our model is able to identify ``\textit{War},'' and ``\textit{conflict}'' as non-triggers based on their context in the sentence, while the baseline seems to be over-emphasizing on their literal meaning.

In some cases, the baseline model also leads to misclassification. For instance,
\textit{The Apache troop \textbf{opened} its tank guns,[...].} Here, ``\textit{opened},'' is an event trigger of type Conflict:Attack, as is correctly identified by our model; however, the baseline misclassifies it as type Movement:Transport.

Another interesting example is: \textit{[...] Beatriz \textbf{walked} into the USCF Offices in New Windsor and immediately \textbf{fired} 17 staff members.} Here, ``\textit{walked}'' is an event trigger of type Movement:Transport, and ``\textit{fired}'' is of type Personnel:End-Position. The baseline model misclassifies ``\textit{fired}'' as Conflict:Attack, while using GTNs help classify it correctly. However, using GTNs can sometimes miss certain event triggers while attempting to be more precise, e.g., ``\textit{walked}'' is missed when using GTNs while the baseline model identifies it correctly.

Lastly, there are examples where both the baseline and proposed models make the same errors.  E.g., \textit{I \textbf{visited} all their families.} or, \textit{I would have \textbf{shot} the insurgent too.} Here, both models misclassify ``\textit{visited},'' (type Contact:Meet) as Movement:Transport, and  ``\textit{shot},'' (type Life:Die) as Conflict:Attack. As future work, we are examining alternate techniques that better inform the context of the event trigger in such sentences. Another interesting example is: ``\textit{It is legal, and \textbf{it} is done.}'' Both models miss ``\textit{it},'' (type Transaction:Transfer-Money). For this example (and some other similar examples of anaphora resolution), we believe that it might be quite non-intuitive to classify the event trigger from the sentence alone, and dependencies among sentences from the same article might need to be leveraged to better inform the context, as we will examine in future work.

\section{Conclusion}

We developed a novel method of enabling existing event extraction models to leverage dependency relations without a significant rise in the number of parameters to be learnt. Our method relies on GTN, and demonstrates an improvement in F1 score over two strong baseline models that do not leverage dependency relations. The benefits of using GTN in an NLP task suggests that other NLP tasks could be improved in the future.

\bibliography{eacl2021}
\bibliographystyle{acl_natbib}

\clearpage
\appendix
\onecolumn
\section{More Details on the MOGANED model} 

There are some subtle differences in the graph-attention mechanisms of Model I and II. In particular, for Model II, the gated-importance $G_{\bm{A}}^0(u,v)$ in equation (1) is redefined as follows: $G_{\bm{A}}^0(u,v){=}\text{softmax}(E(u,v)),$ where $E(u,v){=}A(u,v)\gamma(\bm{W}_{c,\bm{A}}[\bm{W}_{att,\bm{A}}h_u^0||\bm{W}_{att,\bm{A}}h_v^0])$, $\gamma$ is LeakyReLU (with negative input slope $\alpha$), and $\bm{W}_{c,\bm{A}}$ and $\bm{W}_{att,\bm{A}}$ are weight matrices. Further details are provided in \citet{moganed}.

\section{Data Preprocessing} We use the same data split as several existing works~\cite{jrnn,gcnAAAI,liu-etal-2018-jointly,sdp,moganed,relationaware,ji-grishman-2008-refining,liao2010using,li2013joint}, where the sentences from $529$, $30$, and $40$ documents are used as the training, validation, and test set. For preprocessing, we directly used the following code\footnote{\url{https://github.com/nlpcl-lab/ace2005-preprocessing}} which uses the Stanford Core NLP toolkit \cite{manning-etal-2014-stanford}.

\section{Hyper Parameter Setting} For both the models, we select $100$ as the dimension of the word embeddings, and $50$ as the dimension of all the other embeddings, i.e., POS-tag embedding, Named-Entity-type embedding, and positional embedding. Following prior work, we restrict the length of each sentence to be $50$ (truncating long sentences if necessary). We select the hidden units of the BiLSTM network as $100$. We choose a batch size of $10$, and Adam with initial learning rate of $0.0002$. We select the dimension of the graph representation to be $150$. When using GTNs, the number of edge-types ($L$) is $35$, which is determined by the number of unique types of dependency relations, e.g., \verb|nsubj|, \verb|case|, etc., as obtained from the dependency parser.



\end{document}